\documentclass[conference]{IEEEtran}
\IEEEoverridecommandlockouts

\usepackage{cite}
\usepackage{amsmath,amssymb,amsfonts}
\usepackage{algorithm}
\usepackage{algorithmic}
\usepackage{graphicx}
\usepackage{textcomp}
\usepackage{xcolor}
\usepackage{booktabs} 
\usepackage{xurl}      
\usepackage{hyperref} 
\usepackage[T1]{fontenc} 
\usepackage{enumitem}
\usepackage{float}
\usepackage{bibentry}

\def\BibTeX{{\rm B\kern-.05em{\sc i\kern-.025em b}\kern-.08em
    T\kern-.1667em\lower.7ex\hbox{E}\kern-.125emX}}

\newcommand{\etal}{\textit{et al.}}

\begin{document}

\title{Retrieval Augmented Generation (RAG) for Fintech: Agentic Design and Evaluation}

\author{
\IEEEauthorblockN{$^{1,3}$Thomas Cook$^\ast$\thanks{*Work done during internship at Mastercard, Ireland}, $^{2,3}$Richard Osuagwu$^\ast$, 
$^{1,3}$Liman Tsatiashvili$^\ast$,
$^{4,3}$Vrynsia Vrynsia$^\ast$,\\
$^{3}$Koustav Ghosal, $^{3}$Maraim Masoud, $^3$Riccardo Mattivi}
\\
\IEEEauthorblockA{$^1$TU Dublin, Dublin, Ireland\\
$^2$Maynooth University, Ireland\\
$^3$Mastercard, Ireland\\
$^4$National College of Ireland, Ireland\\
\{firstname.lastname\}@mastercard.com}\\
}

\maketitle

\begin{abstract}
Retrieval-Augmented Generation (RAG) systems often face limitations in specialized domains such as fintech, where domain-specific ontologies, dense terminology, and acronyms complicate effective retrieval and synthesis. This paper introduces an agentic RAG architecture designed to address these challenges through a modular pipeline of specialised agents. The proposed system supports intelligent query reformulation, iterative sub-query decomposition guided by keyphrase extraction, contextual acronym resolution, and cross-encoder-based context re-ranking. We evaluate our approach against a standard RAG baseline using a curated dataset of 85 question–answer–reference triples derived from an enterprise fintech knowledge base. Experimental results demonstrate that the agentic RAG system outperforms the baseline in retrieval precision and relevance, albeit with increased latency. These findings suggest that structured, multi-agent methodologies offer a promising direction for enhancing retrieval robustness in complex, domain-specific settings.


\end{abstract}

\begin{IEEEkeywords}
Retrieval Augmented Generation, Agentic AI, Fintech, Natural Language Processing, Knowledge Base, Domain-Specific Ontology, Query Understanding
\end{IEEEkeywords}
\section{Introduction}
\label{sec:intro}
\begin{figure}[htbp]
    \centering
    \includegraphics[width=0.48\textwidth]{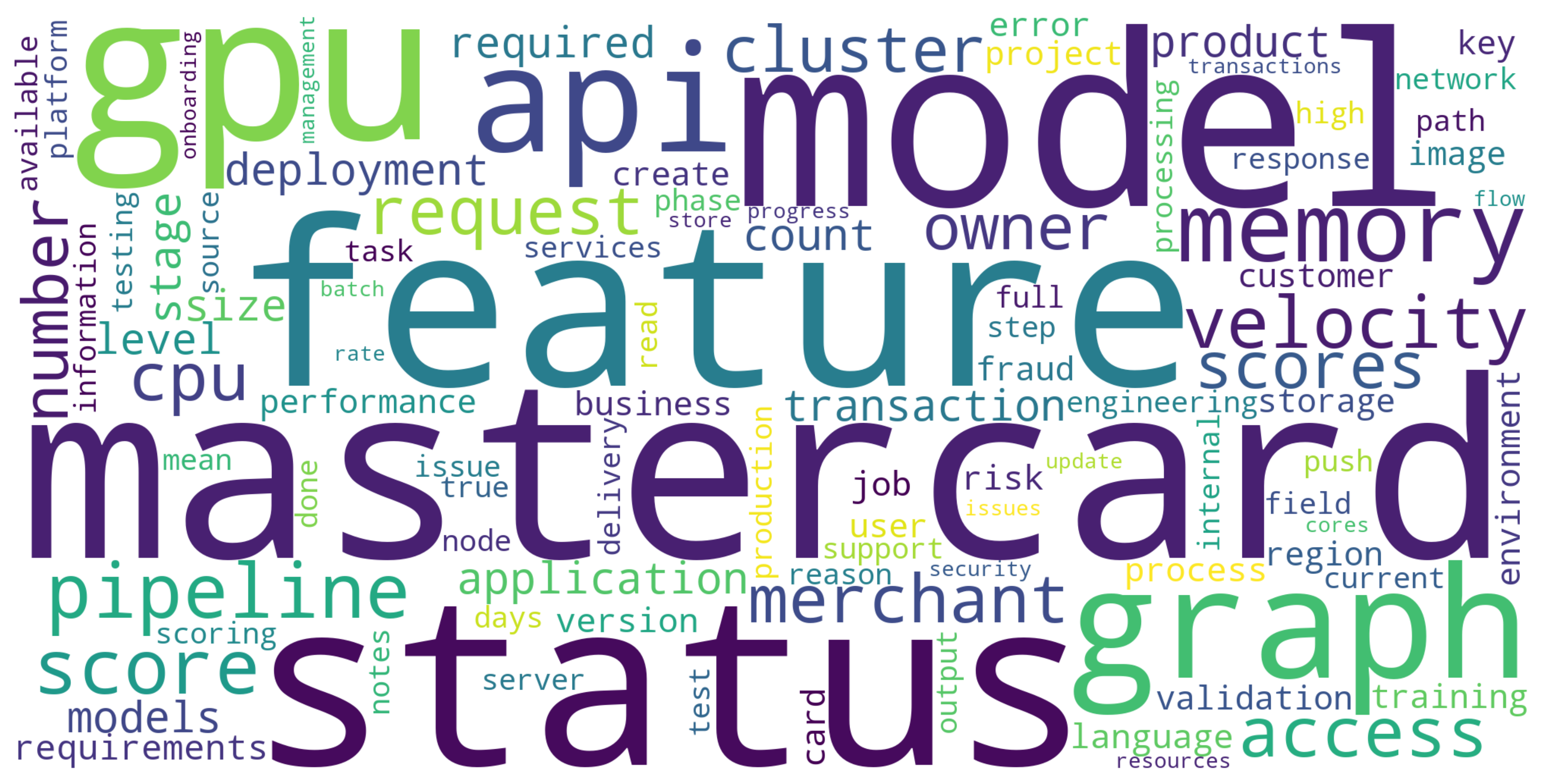}
    \caption{Word cloud illustrating the distribution of internal knowledge artifacts in fintech. The prominence of terms such as “feature,” “status,” “model,” and “API” reflects the operational focus of internal documentation—often centered around technical specifications, product state, and integration interfaces. This concentration of tightly scoped, semi-structured information highlights the challenge of designing RAG systems that can interpret fragmented context across teams and tools, where standard SaaS-based approaches fall short due to regulatory and organizational constraints.}
    \label{fig:wordcloud}
\end{figure}
Retrieval Augmented Generation (RAG) combines large language models (LLMs) with external document retrieval, allowing models to access information beyond their training data. These systems have shown impressive results in general-purpose applications such as technical support \cite{scotti2024llm}, coding assistants \cite{cursor2024}, and document summarization \cite{openai2022chatgpt}, especially when operating as proof-of-concepts on publicly available or loosely structured data.
However, deploying such systems at scale in highly specialized and tightly regulated domains such as financial technology is far from straightforward \cite{schuman2025ai_pilots,menshawy2024navigating}.

Fintech-specific use cases often involve \textit{structured-unstructured data} (or tightly regulated free text), such as proprietary knowledge bases, role-based access to information, and stringent compliance constraints~\cite{10.3389/frai.2018.00001}, all of which make it difficult to repurpose existing tools and methodologies designed for open-domain settings or Software-as-a-Service (SaaS)-based environments. These complexities are both technical and organizational. Regulatory restrictions prohibit data from leaving organisational boundaries, making cloud-hosted APIs or third-party evaluation platforms unsuitable for most practical deployments. In addition, often fintech firms follow a compartmentalized structure with clear divisions of responsibility between product managers, analysts, compliance officers, and engineers. This further complicates how knowledge is captured, retrieved, and applied.

For example, consider a product manager in Mastercard  analyzing feature alignment across offerings such as MDES (Mastercard Digital Enablement Service) 
\footnote{\url{https://www.mastercard.com/newsroom/press/news-briefs/mastercard-digital-enablement-service-mdes}, Accessed: 2025-06-25}, M4M (Mastercard for Merchants) 
\cite{mastercard2020m4m}, and SwitchCore (Mastercard's transaction switching platform) 
\cite{mastercard_switchcore}.
Internal knowledge sources—ranging from note-taking apps, product management platforms and architecture decks to regulatory compliance PDFs—often use acronyms inconsistently or assume implicit context based on team-specific language. This forces the user to manually piece together fragmented information to understand interoperability constraints, roadmap alignment, or API-level integration guidelines. A standard RAG system typically underperforms in such scenarios, frequently misinterpreting short forms like “CMA” (which could mean “Consumer Management Application” or “Cardholder Management Architecture” depending on the context), or retrieving documents based on keyword overlap rather than intent.

Another challenge with such domain-specific use-cases for large enterprises is evaluation. Traditional RAG benchmarks assume public datasets, crowd-sourced relevance judgments, or static ground truth—all of which are infeasible in fintech. Finding subject-matter experts to annotate queries at scale is both costly and time-consuming. Moreover, regulations around confidentiality and data residency prevent the use of crowd platforms for human evaluation.

Therefore, designing on-prem, domain-specific RAG applications requires several considerations and not limited to --- a careful rethinking of architecture, privacy guarantees, a structured retrieval and reasoning process capable of adapting to ambiguity (especially when the source content is internal, fragmented, and often acronym-heavy). In this work, we focus on some of these core challenges and propose an agent design tailored to domain-specific retrieval and reasoning. We also present evaluation methodologies that are secure, reproducible, and feasible at scale within enterprise constraints.

We propose a hierarchical agent framework, where an \textit{Orchestrator Agent} delegates tasks to specialized sub-agents responsible for acronym resolution, domain-aware query rewriting, and cross-encoder-based re-ranking. This modular structure allows the system to encode organizational idiosyncrasies such as internal taxonomies, access controls, and documentation practices, significantly improving the quality of answers retrieved. In addition, we develop a semi-automated evaluation strategy that combines LLM-as-a-judge \cite{liu2023gevalnlgevaluationusing} paradigms with constrained prompt templates and human-in-the-loop validation. Inspired by recent work on synthetic evaluation datasets \cite{liu2023gevalnlgevaluationusing, chen2024benchmarking}, we leverage internal knowledge bases to generate realistic query-answer pairs, which are then assessed using a mix of LLM scoring and manual spot-checking. This allows for a secure, efficient, and consistent evaluation process that scales across teams and data silos.\\

To summarize,  our primary contributions include: 
\begin{itemize}
\item \textbf{A Fintech-focused Agentic Design:} We develop an \textit{Orchestrator Agent} that coordinates specialized agents for acronym resolution, iterative sub-query generation using key phrases, and context refinement through a cross-encoder re-ranking method.
\item \textbf{Automatic Query Enhancement with Continuous Feedback:} Our system proactively identifies and resolves domain-specific acronyms within user queries and retrieved content, enhancing retrieval accuracy and answer completeness.
\item \textbf{Thorough Evaluation Methodology:} We construct an evaluation dataset from an enterprise knowledge base, leveraging LLM-driven synthetic data generation and manual validation to ensure high-quality, contextually relevant questions.
\item \textbf{Empirical Comparative Analysis:} We quantitatively and qualitatively compare our agentic approach against a standard RAG baseline, measuring effectiveness through metrics such as retrieval accuracy and answer relevance.
\end{itemize}

The remainder of this paper is organized as follows. Section~\ref{sec:related_work} reviews relevant work in RAG, multi-agent pipelines, and domain-specific retrieval systems. Section~\ref{sec:agent_design} describes our methodology, including the design of the baseline and agentic RAG architectures. Section~\ref{sec:data_and_evaluation} details the knowledge base preparation and the construction of the evaluation dataset. Section~\ref{sec:experimentation_and_resutls} presents the experimental setup and results, comparing retrieval accuracy and latency across configurations. Section~\ref{sec:discussion} provides a qualitative error analysis and discussion of observed performance patterns. Finally, Section~\ref{sec:conclusion_fucture_work} concludes the paper and outlines directions for future work.
\section{Related Work}
\label{sec:related_work}
\subsubsection*{\textbf{Agentic RAG}}
Recent work has increasingly explored enhancing Retrieval-Augmented Generation (RAG) systems with structured, multi-agent architectures for complex information tasks. Agentic RAG integrates autonomous AI agents into the pipeline, enabling dynamic query decomposition and iterative reasoning. Singh \etal{} introduce the concept of Agentic RAG, where agents perform planning, reflection, and tool use to refine context and retrieval \cite{Singh2025}. Similarly, Nguyen \etal{} propose MA-RAG, a modular, training-free framework where agents (e.g., Planner, Extractor, QA) collaboratively process questions using chain-of-thought reasoning \cite{Nguyen2025}. Their results show that this structure improves answer quality and resolves ambiguities without fine-tuning. The Pathway team further observes that domain-adaptive agents enhance performance by assigning each agent task-relevant expertise \cite{Pathway2025, Nguyen2025}.

\subsubsection*{\textbf{Domain-specific RAG}}
Deploying RAG in specialized domains such as finance, healthcare, and law requires grounding retrieval in domain knowledge. Recent studies emphasize using structured ontologies to enhance LLM performance. Barron \etal{} introduce the SMART-SLIC framework, combining knowledge graphs (KGs) and vector stores customized for a domain to support retrieval \cite{Barron2024}. They demonstrate that referencing KG ontologies improves QA by aligning retrieval with relevant subgraph structures. Their approach fuses structured (KG) and unstructured (text) sources to reduce hallucinations and improve citation fidelity \cite{Barron2024}. However, they also highlight the resource demands of curating such domain-specific infrastructure. Other studies similarly confirm that structured knowledge sources boost retrieval precision in specialized settings such as legal and medical QA \cite{Barron2024, Singh2025}.

\subsubsection*{\textbf{LLM Applications in Fintech}}
Large language models have found wide adoption in fintech, powering analytics and automation across customer support, fraud detection, risk modeling, and compliance. Daiya surveys how LLMs monitor financial risks by processing vast unstructured data, detecting patterns, and forecasting threats from logs, news, and communications \cite{Daiya2024}. Broughton highlights their role in fraud detection, credit scoring, and automating regulatory tasks like document parsing and report generation \cite{Broughton2024}. These studies agree that LLMs extract actionable insights from diverse financial sources, whether through sentiment analysis, trading signal identification, or customer feedback mining \cite{Broughton2024, Daiya2024}. These applications underscore the need for domain-aware RAG to enhance factual accuracy in high-stakes financial contexts.

\subsubsection*{\textbf{RAG for Fintech}}
Within fintech, RAG systems are being prototyped for data-driven assistants and analytics tools. Lumenova AI reports growing adoption of RAG-backed chatbots that integrate live market and account data to improve customer interaction \cite{Lumenova2024}. A notable example is Bank of America’s “Erica,” which uses LLMs and real-time retrieval to support over a billion user interactions \cite{Lumenova2024}. Hernandez Leal demonstrates a RAG-powered assistant trained on SEC filings to answer investor questions from 10-K and 10-Q reports \cite{Leal2024}. These examples show how RAG can anchor LLM outputs in authoritative financial sources, enhancing both precision and citation. While still in early stages, these systems suggest that real-time, curated retrieval pipelines significantly boost LLM utility and trustworthiness in financial applications \cite{Lumenova2024, Leal2024}.

\section{Agent Design}
\label{sec:agent_design}
Building on the challenges introduced in Section~\ref{sec:intro}, this section details how we first implemented a Baseline RAG pipeline to explore retrieval limitations in fintech settings, and subsequently refined the design into a modular agentic architecture.

\subsection{Baseline RAG System}

\begin{figure}[htbp]
\centering
\includegraphics[height=6cm]{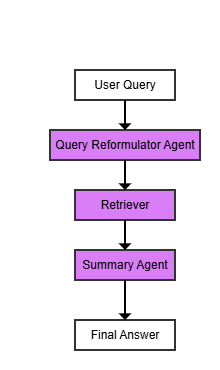}
\caption{Overview of the B-RAG pipeline. The process follows a linear flow beginning with the user's initial query, followed by query reformulation, single-pass retrieval, summarization, and response generation.}
\label{fig:baseline_flow}
\end{figure}

The Baseline RAG system, which will be referred to as "B-RAG" for the remainder of this work, represents a standard retrieval pipeline that serves as a comparison point for our agentic architecture. This approach follows a sequential process, as illustrated in Figure~\ref{fig:baseline_flow}.

The B-RAG system, illustrated in Figure~\ref{fig:baseline_flow}, consists of a sequence of specialized agents structured to enable stepwise information retrieval and synthesis.  The process begins with the user’s initial input, which serves as the contextual foundation for subsequent components. This input is then propagated through the following agents:

\begin{enumerate}[label=(\alph*)]
    \item \textit{Query Reformulator Agent}: Takes the user's question and refines it into a concise, keyword-focused query optimized for retrieval. It uses predefined prompt templates to detect if the query is a continuation of previous interactions or a standalone new topic.
    \item \textit{Retriever Agent}: Executes a single-pass retrieval from the knowledge base with cosine similarity between embeddings to locate relevant document chunks.    
    \item \textit{Summary Agent}: Aggregates retrieved document chunks into a coherent answer. It is explicitly instructed to avoid generating information not present in the retrieved context, clearly indicating if the retrieved content lacks sufficient information.
\end{enumerate}

Finally, the output of the summary agent is  presented  as a coherent response, supported by the reference links.

\noindent The B-RAG pipeline serves as a reference due to its simplified architecture, which allows us to conduct clear and interpretable performance benchmarking against the more complex agentic system. However, this simplicity imposes several notable limitations:

\begin{itemize}
    \item \textit{Lack of multi-pass retrieval logic}: The system performs a single retrieval pass without iterative refinement, thereby constraining its capacity to effectively address complex queries that require deeper exploration.
    \item \textit{Absence of sub-query decomposition}: The pipeline is unable to partition broad or ambiguous queries into multiple targeted sub-queries. This limitation results into diminished precision when questions span multiple document fragments.

    \item \textit{No acronym resolution mechanism}: The system does not resolve domain-specific acronyms frequently encountered in enterprise context, which may adversely affect the clarity and accuracy of retrieved information.
    
    \item \textit{Absence of document re-ranking}: Retrieved results are utilized without further subsequent validation or refinement through cross-encoder-based re-ranking. This potentially may compromise the relevance of the final responses.

\end{itemize}

Despite these limitations, the simple design of the baseline pipeline offers some advantages, including ease of deployment, lower computational overhead, and enhanced interpretability. These attributes make it an effective benchmark for systematically evaluating the incremental improvements introduced by each advanced component of the proposed agentic architecture.

\subsection{Proposed Refined Design}

The Agentic RAG system, which will be referred to as "A-RAG" throughout this paper,  is designed to address the limitations inherent in the B-RAG system by integrating modular intelligence and task-specific specialization. Central to A-RAG is the \textit{Orchestrator Agent}, which coordinates a suite  of specialized agents, each tasked with a distinct retrieval or synthesis function. This design is inspired by an investigative research workflow. It initiates with a direct attempt to address the user query, subsequently evaluates the output, and when necessary, activates more focused and iterative exploration.

\begin{figure}[htbp]
    \centering
    \includegraphics[width=0.48\textwidth]{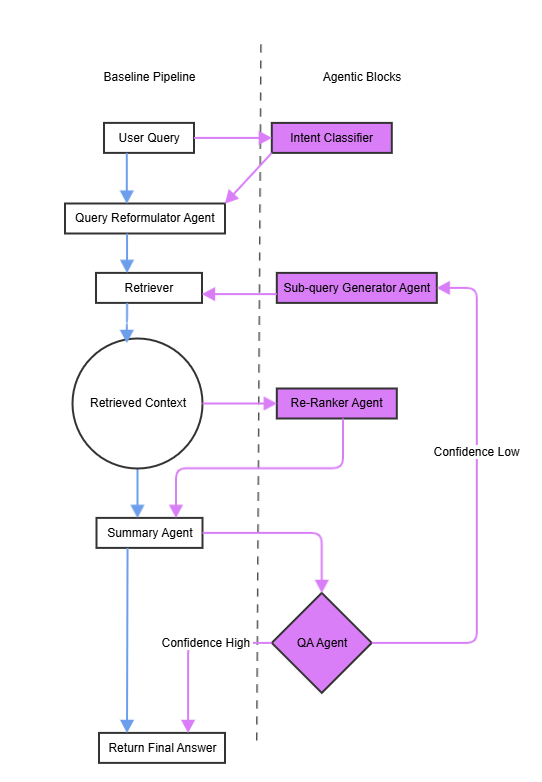}
    \caption{Comparison of hybrid pipeline architectures for B-RAG and A-RAG workflows. The left panel shows B-RAG’s single-pass process, including query reformulation, retrieval, answer synthesis, and output generation without iterative refinement. The right panel illustrates A-RAG’s extended pipeline, featuring acronym resolution, sub-query generation, document re-ranking, and an answer quality assessment (QA) agent. If the QA agent assigns low confidence to the synthesized answer (e.g., below a set threshold), a feedback loop triggers sub-query generation to iteratively expand and refine retrieval.}
    \label{fig:comparison}
\end{figure}

To address challenges such as ambiguous terminology, fragmented sources, and answer confidence, A-RAG incorporates several key capabilities, including acronym resolution, sub-query generation, parallel retrieval, re-ranking, and quality assessment. The quality of generated answers is evaluated by a dedicated QA agent whose confidence score determines whether further iterative refinement is required. A comparison of the hybrid pipeline architectures for B-RAG and A-RAG workflows is shown in Figure~\ref{fig:comparison}. It highlights the differences in processing flow and iterative mechanisms.

\subsection{System Architecture and Workflow}

The A-RAG system comprises a set of lightweight, modular agents orchestrated to support iterative retrieval and reasoning. The following subsections describe the system's core agent components and the operational workflow using a representative fintech query example.

\subsubsection*{Core Agent Components}

The orchestrator coordinates eight specialized agents, each responsible for a distinct stage in the retrieval–reasoning cycle:

\begin{enumerate}[leftmargin=*, noitemsep]

    \item \textbf{Intent Classifier} – determines whether the user input requires  
          \textit{(i) retrieval}, which involves fetching new context from the knowledge base, or  
          \textit{(ii) summary}, which compresses existing conversational history.

    \item \textbf{Query Reformulator} – transforms the raw query into a dense, keyword-optimized search string by removing function words, expanding recognized acronyms, and injecting domain-specific synonyms to maximize embedding-based recall.

    \item \textbf{Retriever Manager \& Retriever Agent} – launch one or more vector store queries—executed in parallel if sub-queries exist—and return the top-$k$ most relevant chunks.

    \item \textbf{Sub-Query Generator} – in cases of low initial retrieval scores, identifies 2–3 key entities or phrases to construct targeted follow-up queries.

    \item \textbf{Re-Ranker Agent} – reorders the retrieved chunks using a cross-encoder to prioritize those with higher semantic alignment to the query.

    \item \textbf{Summary Agent} – fuses the ranked context snippets into a concise, citation-style answer composed strictly from the retrieved content.

    \item \textbf{QA Agent} – evaluates the synthesized answer on a 0–10 scale, assessing relevance and support from context; the score determines whether the pipeline concludes or proceeds with refinement.

    \item \textbf{Acronym Resolution Logic} – manages a local glossary and, when invoked, appends in-line definitions to prevent downstream agents from misinterpreting shorthand expressions.
\end{enumerate}

\subsubsection*{Workflow Overview}







The operational workflow integrates the agent components into a cohesive, adaptive pipeline, illustrated with the example query: \textit{“How is CVaR calculated in the IRRBB framework?”} The process begins with query reformulation and acronym expansion, ensuring clarity and disambiguation of terms like \texttt{CVaR} and \texttt{IRRBB}. A first-pass vector retrieval fetches relevant document chunks, which are synthesized into an initial answer. If the QA agent assigns a low confidence score, the system triggers refinement: sub-queries such as “CVaR formula” or “IRRBB risk quantification” are generated, results are re-ranked, and a revised synthesis is attempted. Should this prove inadequate, a broader retrieval sweep is conducted to increase coverage. In the absence of a sufficiently confident answer, the system transparently communicates its uncertainty to the user. This adaptive pipeline dynamically balances computational efficiency with retrieval depth based on answer confidence.

\section{Data and Evaluation Strategy}
\label{sec:data_and_evaluation}
\label{sec:data}

In Section~\ref{sec:intro}, we described the challenge of designing retrieval systems that can operate effectively over fragmented and acronym-heavy enterprise knowledge bases. This section details how we constructed the knowledge corpus and generated an evaluation set tailored to this context.

\subsection{Knowledge Base Preparation}
The knowledge base was derived from proprietary internal documents exported in structured markup format. A custom pipeline was developed to preprocess the data. This process focused on:
\begin{itemize}
    \item Converting structured elements such as tables and code blocks into linear, plain-text representations while preserving semantic relationships.
    \item Removing formatting noise and markup artifacts to produce clean, model-ready text.
\end{itemize}
To enable retrieval, the documents were segmented into overlapping chunks (to preserve contextual coherence across boundaries), embedded with a publicly available sentence transformer, and stored in a vector index.

\begin{figure}[htbp]
    \centering
    \includegraphics[width=0.48\textwidth]{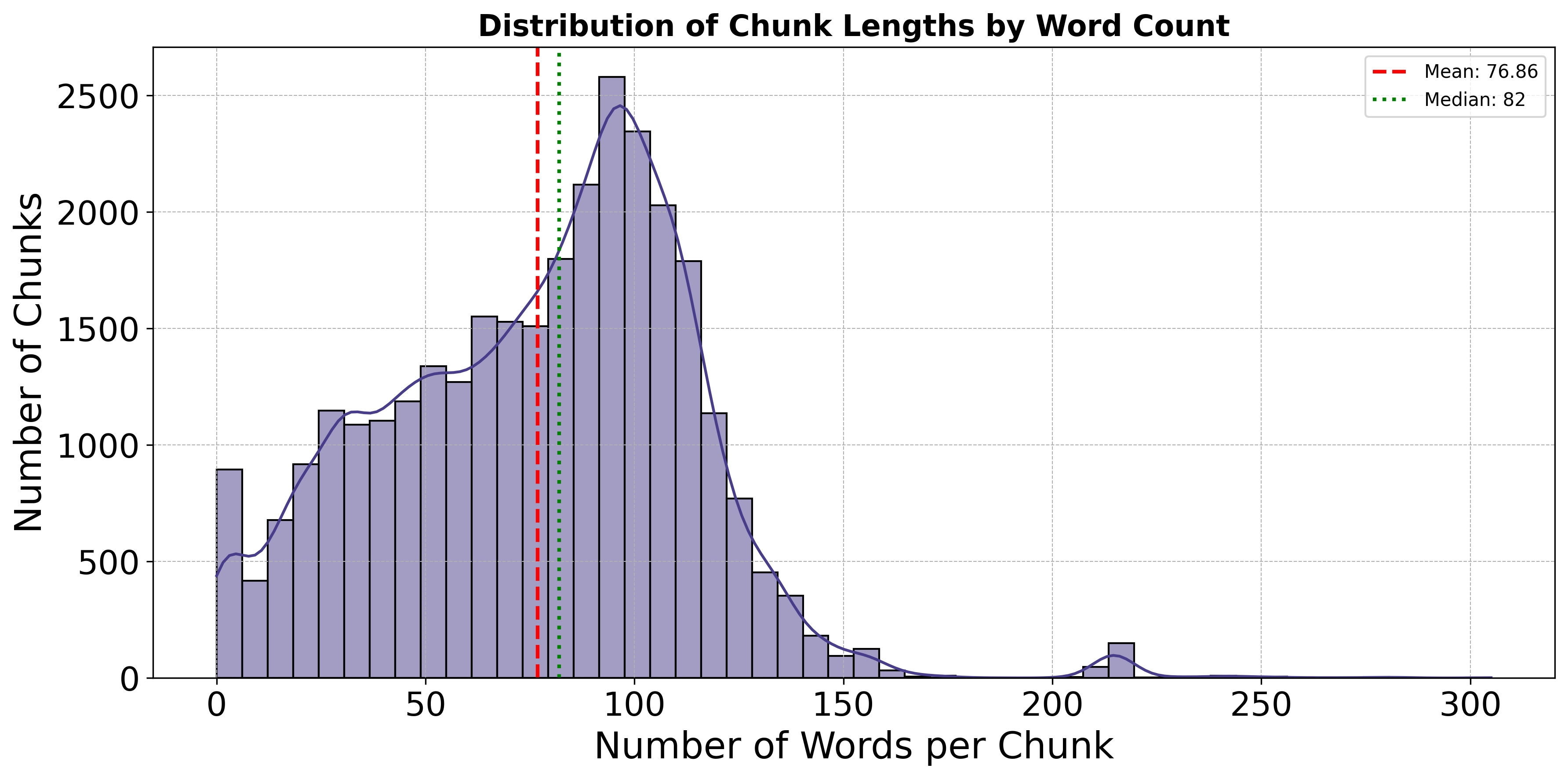}
    \caption{Distribution of chunk lengths by word count. The majority of chunks are between 50 and 120 words, showing the trade-off between retrieval granularity and contextual coherence.}
    \label{fig:chunk_distribution}
\end{figure}

The resulting corpus consisted of over 30,000 text chunks representing 1,624 unique documents. On average, each document was split into approximately 19 chunks. Figure~\ref{fig:chunk_distribution} shows the distribution of chunk lengths measured by word count.

\begin{figure}[htbp]
    \centering
    \includegraphics[width=0.48\textwidth]{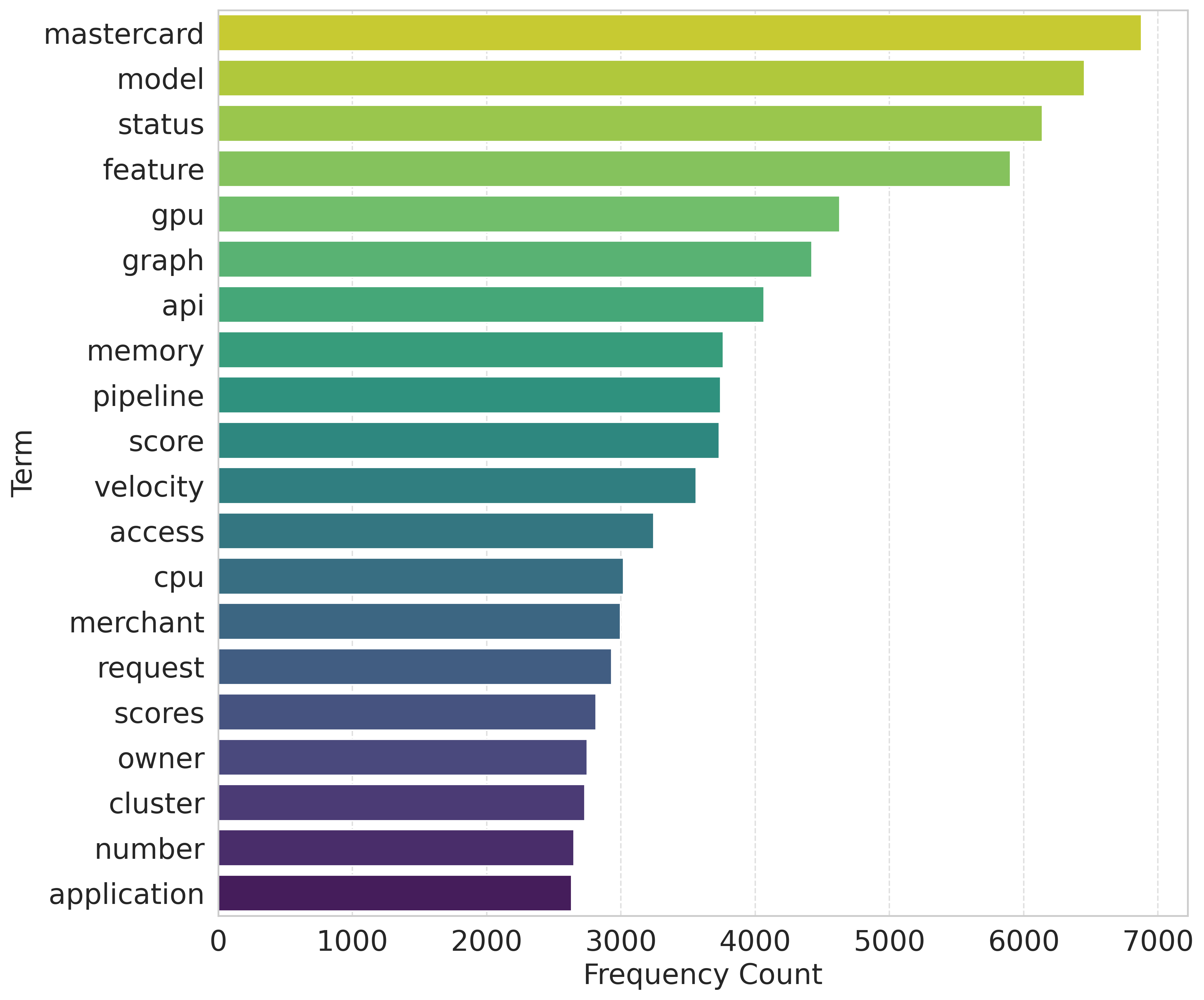}
    \caption{Top 20 most frequent terms in the corpus, ranked by raw frequency.}
    \label{fig:top_terms_barchart}
\end{figure}

To further illustrate the dataset composition, Figure~\ref{fig:wordcloud} and Figure~\ref{fig:top_terms_barchart} provide complementary views of term frequency. Both highlight the prevalence of domain-specific language in the internal fintech corpus, particularly terms such as “feature,” “API,” and “model,” reflecting its emphasis on product, technical, and compliance-related content.


\subsection{Evaluation Set Generation}
Inspired by the LLM-as-a-judge paradigm~\cite{liu2023gevalnlgevaluationusing}, we designed a multi-phase, model-assisted pipeline to create a high-quality evaluation set. This approach is motivated by the constraints outlined in Section~\ref{sec:intro}, particularly the impracticality of relying solely on manual annotation within confidential enterprise environments.

First, we randomly sampled a subset of chunks from the curated corpus. For each chunk, a language model was prompted to generate a question-answer pair grounded in the provided context. Two distinct prompt templates were used; one tailored to narrative text and the other to  structured elements such as tables. This distinction is made to ensure both content diversity and fidelity to the source material.

Subsequently, each generated question–answer pair underwent a quality control phase in which a language model assessed contextual alignment, factual accuracy, and coherence. To standardize this evaluation, we defined three criteria that each pair was required to meet:

\begin{itemize}
    \item \textbf{Specificity:} Is the question clearly and narrowly scoped to a particular aspect of the context?
    \item \textbf{Faithfulness:} Is the answer grounded exclusively in the provided source text, without introducing external information?
    \item \textbf{Completeness:} Does the answer fully and directly address the question posed?
\end{itemize}
Only candidate pairs that satisfied all three criteria were retained. These pairs then underwent an additional round of manual review to eliminate any remaining edge cases or ambiguous examples. Finally, the resulting evaluation set consisted of 85 validated question-answer pairs, each traceable to a unique document chunk. This dataset served as the foundation for the  quantitative and qualitative evaluations presented in Sections~\ref{sec:metrics}, \ref{sec:llm_judge}.

\subsection{Human-Verified Prompt–Answer Benchmark}
\label{sec:dataset_curate}

In addition to the automated evaluation set described above, we constructed a manually curated benchmark designed to stress-test both retrieval fidelity and answer ranking. The benchmark construction process involved the following key steps:
\paragraph*{Question sourcing}Candidate questions were extracted from routine activities within the internal \textit{knowledge base}, including onboarding runbooks, intern reports, and project or product wikis.

\paragraph*{Human validation of answers}For each question, human annotators identified all relevant documentation pages and selected one representative chunk per page as a potential answer. Among these, the most comprehensive chunk was designated as the ground truth. This setup reflects a “one correct, many plausible” retrieval scenario.

\paragraph*{Intent categorization}Each query was labeled according to an intent taxonomy—\textit{Procedural}, \textit{Definitional}, or \textit{Acronym Expansion}—to support stratified analysis, as discussed in Section~\ref{sec:llm_judge}.

\paragraph*{Dataset composition}The final benchmark comprises 27 question–answer pairs, each corresponding to a distinct query with multiple associated ground-truth answers: seven procedural, fourteen definitional, and six acronym-expansion examples.\\

This human-verified dataset was used to (i) compute strict retrieval metrics (\texttt{Hit@k}), and (ii) support error analysis in the presence of multiple valid answers. For each item, we log the following fields: \textit{Question}, \textit{Category}, \textit{Ground-truth Answer(s)}, \textit{Ground-truth Source(s)}, \textit{Generated Answer}, and \textit{Retrieved Source(s)}.

\section{Experiments and Results}
\label{sec:experimentation_and_resutls}
\subsection{Experimental Setup}

We compared B-RAG and A-RAG pipelines on a specialized internal fintech knowledge base. Both pipelines used the same LLM backend, \texttt{Llama-3.1-8B-Instruct}, served via a vLLM endpoint. Document embeddings were generated using the \texttt{all-MiniLM-L6-v2} model and stored in ChromaDB. We evaluated both systems using two key metrics: (1) Retrieval Accuracy (Hit Rate @5), defined as the percentage of questions for which the correct document source appeared among the top five retrieved links; and (2) Average Latency, defined as the time from user query submission to the final system response.

\subsection{Quantitative Results}
\label{sec:metrics}
Table \ref{tab:main_results} summarizes the results. The A-RAG system achieved a strict retrieval accuracy of 62.35\%, outperforming the Baseline's 54.12\%. This performance gain can be attributed to the system’s specialized agents for acronym resolution and sub-query expansion, which were designed to address the fragmented and semantically sparse nature of enterprise knowledge described in Section~\ref{sec:intro}. As expected, A-RAG's average query latency was 5.02 seconds, significantly higher than B-RAG’s 0.79 seconds, due to multi-stage processing and iterative document re-ranking.

\begin{table}[htbp]
    \caption{Performance Comparison of B-RAG and A-RAG Systems}
    \begin{center}
    \begin{tabular}{lcc}
        \toprule
        \textbf{Metric} & \textbf{B-RAG} & \textbf{A-RAG} \\
        \midrule
        Total Questions Evaluated & 85 & 85 \\
        Retrieval Accuracy (Hit @5, \%) & 54.12 & 62.35 \\
        Avg. Latency per Query (s) & 0.79 & 5.02 \\
        \bottomrule
    \end{tabular}
    \label{tab:main_results}
    \end{center}
\end{table}

\paragraph{Adjusted Retrieval Interpretation} Although retrieval was evaluated using exact link matches, manual inspection identified several queries where the system returned correct content from semantically equivalent but non-identical documents. Specifically, A-RAG retrieved valid answers from alternate sources in six additional cases, and B-RAG in three. Incorporating these, the adjusted retrieval accuracy increases to 69.41\% for A-RAG and 58.82\% for B-RAG. These cases reinforce the core challenge introduced in Section~\ref{sec:intro}; that enterprise knowledge in fintech domains is often fragmented, with key information distributed across related documents. \\
Unlike B-RAG, which often fails when the exact match is missing, A-RAG's sub-query generation and iterative re-ranking modules better synthesize partial context across sources, producing correct responses even when the originating chunk is not directly retrieved. In the next section we evaluate answer quality using \emph{semantic accuracy}, defined as the mean LLM-judge score measuring semantic equivalence between a system’s answer and the human ground-truth answer, independent of surface lexical overlap.

\subsection{Answer-Quality Evaluation (Semantic Accuracy)}
\label{sec:llm_judge}
Building on the retrieval metrics in Table~\ref{tab:main_results} (Section~\ref{sec:metrics}), we next measure how faithfully each pipeline answers user questions.  Rather than lexical overlap, we target \emph{semantic} agreement with the ground-truth answers created in Section~\ref{sec:data}.

We define the following metric to calculate the semantic accuracy. Let $N$ denote the total number of evaluated questions (here $N=85$). For every question $q_i$, an external vLLM-hosted judge (\texttt{Llama-3.1-8B}) assigns an integer score $s_i \in \{1,\ldots,10\}$ using the rubric in Table~\ref{tab:llm_judge_rubric}. The semantic-accuracy metric is the mean judge score:
\[
\bar{s} = \frac{1}{N}\sum_{i=1}^{N} s_i.
\]

\begin{table}[htbp]
    \centering
    \caption{LLM-judge rubric (semantic accuracy).}
    \label{tab:llm_judge_rubric}
    \begin{tabular}{cl}
        \toprule
        Score & Interpretation \\ \midrule
        9–10 & Exact match or perfect paraphrase \\
        6–8  & Correct but missing minor detail \\
        3–5  & Honest refusal / incomplete answer \\
        1–2  & Incorrect or hallucinated \\ \bottomrule
    \end{tabular}
\end{table}

Applying this metric to our evaluation set reveals that the A-RAG pipeline  outperforms the baseline B-RAG. A-RAG achieves a mean score of $\bar{s} = 7.04$, compared to $\bar{s} = 6.35$ for B-RAG, yielding a performance gain of $\Delta\bar{s} \approx 0.68$.

\begin{figure}[htbp]
    \centering
    \includegraphics[width=0.45\textwidth]{"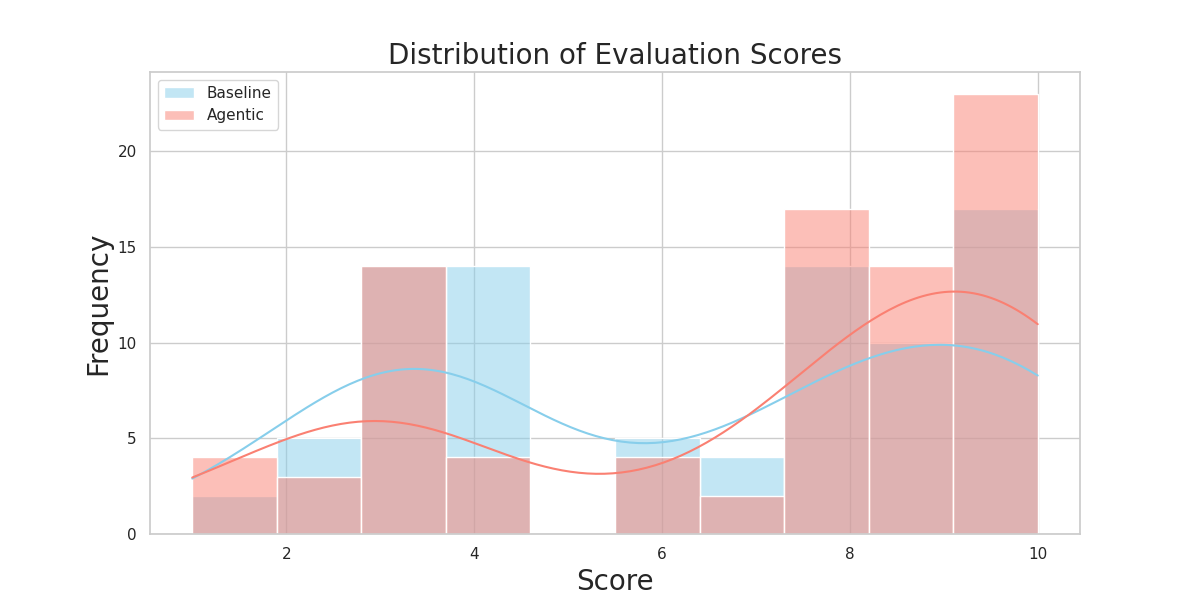"}
    \caption{Histogram+KDE of all 85 LLM-judge scores per pipeline. A-RAG shifts mass into the 7–10 band, halves low-quality answers ($s<5$ falls from 18\% to 8\%), and increases “excellent” ($s\ge9$) responses from 12\% to 22\%.}
    \label{fig:llm_judge_dist}
\end{figure}

\begin{figure}[H]
    \centering
    \includegraphics[width=0.45\textwidth]{"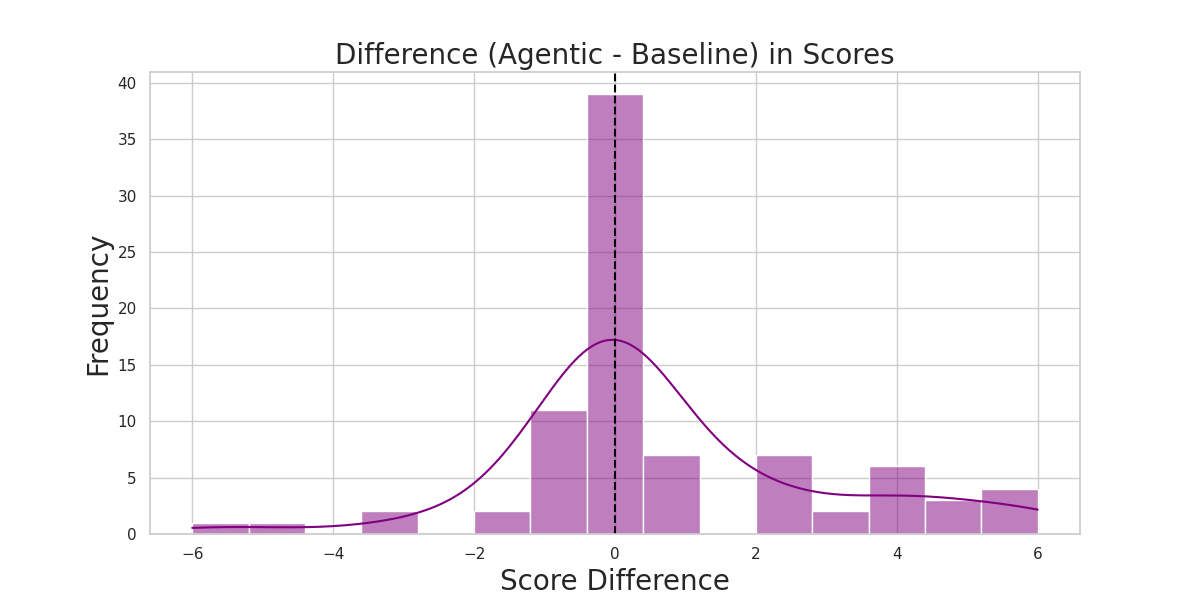"}
    \caption{Per-question score difference $\Delta s=s_{\text{A-RAG}}-s_{\text{B-RAG}}$.  Distribution centers at $+1$ (median) with 64\% of points $>0$, 25\% at 0, and only 11\% $<0$; extreme negative swings never exceed $-3$, indicating A-RAG rarely degrades answer quality.}
    \label{fig:llm_judge_diff}
\end{figure}

As shown in Figure~\ref{fig:llm_judge_dist}, A-RAG notably reduces the frequency of low-quality answers ($s{<}5$) and increases the proportion of responses in the top rubric tier ($s \geq 9$). The per-question score deltas (Figure~\ref{fig:llm_judge_diff}) show that A-RAG is preferred in 64\,\%  of cases, ties in 25\,\%  and is outperformed by B-RAG in only 11\,\%. Even in those rare cases, the largest drop is just 3 points, indicating that A-RAG rarely degrades answer quality.




\subsection{Human-Curated Benchmark}
We further evaluate both systems using a human-curated benchmark consisting of 17 questions, each designed to allow multiple plausible answers. The benchmark comprises 9 definitional questions, 4 procedural questions, and 4 acronym-based questions. Collectively, these questions are associated with 33 distinct ground-truth source links. In addition to assessing semantic accuracy, we introduce coverage as a complementary metric:
\[
\mathrm{Coverage}=\frac{|\mathcal{R}|}{|\mathcal{G}|},
\]
where $\mathcal{G}$ is the set of 33 distinct ground-truth source links and $\mathcal{R}$ the subset retrieved within the top–5 across all questions. B-RAG achieves 66.67\% (22/33), while A-RAG attains 69.70\% (23/33). This result indicates a modest advantage in aggregating distributed evidence.

\begin{table}[htbp]
\centering
\caption{Human-curated benchmark results (Coverage and Semantic Accuracy)}
\label{tab:human_benchmark}
\begin{tabular}{lcccc}
\toprule
System & Category & \#Questions & Coverage (\%) & Semantic Acc. \\
\midrule
B-RAG & Overall      & 17 & 66.67 & 7.88 \\
      & Definitional & 9 & 73.68  &  7.78\\
      & Procedural   & 4  &  57.14  &  7.75\\
      & Acronym      & 4  &  57.14  & 8.25\\
A-RAG & Overall      & 17 & 69.70 &  8.06\\
      & Definitional & 9 &  68.42  &  7.89\\
      & Procedural   & 4  &  100.0 &  8.25\\
      & Acronym      & 4  &  42.85  &  8.25\\
\bottomrule
\end{tabular}
\end{table}

\section{Discussion}
\label{sec:discussion}
\subsection{Interpretation of Results}
The results indicate that the A-RAG system achieved measurable improvements in retrieval accuracy over the Baseline, confirming our hypothesis that agentic decomposition strategies are beneficial in fragmented, domain-specific settings. The strict accuracy gain (62.35\% vs. 54.12\%) is further supported by manual review, where A-RAG often retrieved semantically correct information from alternative sources not marked as ground truth. These observations align with the core challenge described in Section I: enterprise fintech documents are semantically sparse and often distribute relevant information across multiple partially redundant pages.

Qualitative analysis revealed that the most effective agentic component was sub-query generation, which enabled targeted exploration of edge cases where the initial query lacked specificity. Acronym resolution proved more error-prone: in cases of ambiguous or undefined acronyms, the agent occasionally surfaced overly generic sources. The cross-encoder re-ranking improved relevance when multiple near-matches were retrieved but introduced latency. These trade-offs confirm that while A-RAG design introduces computational overhead, its modular structure is more robust to domain-specific ambiguity than the static B-RAG pipeline. 

As shown in the human-curated benchmark (Table~\ref{tab:human_benchmark}), A-RAG achieves higher overall coverage (69.70\% vs.\ 66.67\%) and a modest improvement in semantic accuracy (8.06 vs.\ 7.88), indicating enhanced consolidation of dispersed evidence. The gains are most evident in procedural queries, where coverage reaches 100\% and semantic quality rises to 8.25. This suggests that iterative sub-query expansion is particularly effective when task steps are implicit. In contrast, acronym and definitional queries do not show consistent improvements—acronym coverage declines (57.14\% to 42.85\%), and definitional coverage decreases slightly. These outcomes imply that acronym resolution and re-ranking mechanisms may occasionally over-filter or misprioritize near-duplicate sources. Nevertheless, the identical acronym semantic scores (8.25) suggest that A-RAG can synthesize sufficient context from partial retrieval. Overall, the “many plausible answers” setting underscores both the robustness advantages of agentic orchestration and specific limitations (e.g., acronym recall) that warrant further investigation.


\subsection{Addressing Domain-Specific Ontology}
The A-RAG pipeline explicitly handled two challenges central to fintech ontologies: acronym disambiguation and context fragmentation. By integrating an acronym helper agent and embedding definitions where available, the system was able to expand domain-specific abbreviations during both query reformulation and document re-ranking. However, this component showed limitations in edge cases where acronyms were neither defined nor contextually grounded in the document. In contrast, B-RAG relied entirely on surface-level keyword overlap, often missing results that required any interpretive inference.

Sub-query phrasing also improved navigation through the ontology by decomposing vague or overloaded user queries into focused, domain-aligned sub-tasks. This was  impactful for queries that implicitly referenced process hierarchies or tools—common in enterprise documentation. Overall, the A-RAG system demonstrated a stronger ability to operate within the constraints of a specialized ontology, especially when semantic cues in the documents were minimal or inconsistently structured.

\subsection{Limitations of the Current Study}
The study is constrained by several factors. Firstly, the evaluation dataset is relatively small (85 questions) and may not reflect the full spectrum of real-world queries or document structures in fintech. Secondly, all results are grounded in the use of a single LLM (\texttt{Llama-3.1-8B-Instruct}) and embedding model, limiting generalization across architectures. Thirdly, certain agents (e.g., acronym resolver) rely on heuristic definitions and simple regex-based expansion, which may under perform in more complex acronym usage. Lastly, while adjusted accuracy was discussed, it is not backed by a formal metric; a more rigorous semantic equivalence measure should be incorporated in future work to better capture retrieval performance in fragmented corpora.

\section{Conclusion and Future Work}
\label{sec:conclusion_fucture_work}

This study investigated whether a modular, agent-driven RAG pipeline could more effectively navigate fragmented enterprise knowledge bases in the fintech domain. The proposed A-RAG system incorporated acronym resolution, subquery decomposition, and document reranking agents to enable more adaptive retrieval strategies, yielding a measurable improvement in strict retrieval accuracy (62.35\% vs. 54.12\%) over a standard baseline, and rising to 69.41\% when accounting for semantically relevant but non-ground-truth sources. 

Looking ahead, future work could explore more principled design strategies for agent coordination, such as reinforcement learning or meta-controller frameworks that dynamically adapt the agent composition based on query type or retrieval feedback. Additionally, integrating stronger context-awareness—such as discourse-level tracking, temporal grounding, or multi-turn memory could help align answers more closely with user intent. Enhancing the reasoning capabilities of the pipeline remains an open challenge, particularly in producing not just faithful but helpful responses that satisfy nuanced informational needs. Techniques such as agent self-critique, counterfactual retrieval, or reflection-based loops may offer promising paths forward. We expect this work contribute a meaningful step toward more robust and interpretable retrieval-augmented systems in high-stakes, domain-specific applications.


\bibliographystyle{IEEEtran}
\bibliography{conference_101719} 

\begin{thebibliography}{10}
\providecommand{\url}[1]{#1}
\csname url@samestyle\endcsname
\providecommand{\newblock}{\relax}
\providecommand{\bibinfo}[2]{#2}
\providecommand{\BIBentrySTDinterwordspacing}{\spaceskip=0pt\relax}
\providecommand{\BIBentryALTinterwordstretchfactor}{4}
\providecommand{\BIBentryALTinterwordspacing}{\spaceskip=\fontdimen2\font plus
\BIBentryALTinterwordstretchfactor\fontdimen3\font minus \fontdimen4\font\relax}
\providecommand{\BIBforeignlanguage}[2]{{%
\expandafter\ifx\csname l@#1\endcsname\relax
\typeout{** WARNING: IEEEtran.bst: No hyphenation pattern has been}%
\typeout{** loaded for the language `#1'. Using the pattern for}%
\typeout{** the default language instead.}%
\else
\language=\csname l@#1\endcsname
\fi
#2}}
\providecommand{\BIBdecl}{\relax}
\BIBdecl

\bibitem{scotti2024llm}
V.~Scotti and M.~J. Carman, ``Llm support for real-time technical assistance,'' in \emph{Joint European Conference on Machine Learning and Knowledge Discovery in Databases}.\hskip 1em plus 0.5em minus 0.4em\relax Springer, 2024, pp. 388--393.

\bibitem{cursor2024}
{Cursor AI}, ``Cursor: The ai-first code editor,'' \url{https://www.cursor.so}, 2024, accessed: 2025-06-25.

\bibitem{openai2022chatgpt}
{OpenAI}, ``Chatgpt: Optimizing language models for dialogue,'' \url{https://openai.com/blog/chatgpt}, 2022, accessed: 2025-06-25.

\bibitem{schuman2025ai_pilots}
E.~Schuman, ``{88\% of AI pilots fail to reach production — but that's not all on IT},'' \url{https://www.cio.com/article/3850763/88-of-ai-pilots-fail-to-reach-production-but-thats-not-all-on-it.html}, Mar. 2025, accessed: 2025-06-25.

\bibitem{menshawy2024navigating}
A.~Menshawy, Z.~Nawaz, and M.~Fahmy, ``Navigating challenges and technical debt in large language models deployment,'' in \emph{Proceedings of the 4th Workshop on Machine Learning and Systems}, 2024, pp. 192--199.

\bibitem{10.3389/frai.2018.00001}
\BIBentryALTinterwordspacing
P.~Giudici, ``Fintech risk management: A research challenge for artificial intelligence in finance,'' \emph{Frontiers in Artificial Intelligence}, vol. Volume 1 - 2018, 2018. [Online]. Available: \url{https://www.frontiersin.org/journals/artificial-intelligence/articles/10.3389/frai.2018.00001}
\BIBentrySTDinterwordspacing

\bibitem{mastercard2020m4m}
{Mastercard}, ``Mdes for merchants (m4m): Mastercard digital enablement service for merchants,'' \url{https://www.mastercard.com/news/eemea/en/newsroom/press-releases/en/2020/august/mastercard-research-shows-surge-in-digital-payments-as-ecommerce-reaches-new-heights-in-the-uae/}, 2020, accessed: 2025-06-25.

\bibitem{mastercard_switchcore}
------, ``Switchcore: Mastercard transaction switching platform,'' \url{https://www.mastercard.com/eea/switching-services/our-technology/network.html}, 2025, accessed: 2025-06-25.

\bibitem{liu2023gevalnlgevaluationusing}
\BIBentryALTinterwordspacing
Y.~Liu, D.~Iter, Y.~Xu, S.~Wang, R.~Xu, and C.~Zhu, ``G-eval: Nlg evaluation using gpt-4 with better human alignment,'' 2023. [Online]. Available: \url{https://arxiv.org/abs/2303.16634}
\BIBentrySTDinterwordspacing

\bibitem{chen2024benchmarking}
J.~Chen, H.~Lin, X.~Han, and L.~Sun, ``Benchmarking large language models in retrieval-augmented generation,'' in \emph{Proceedings of the AAAI Conference on Artificial Intelligence}, vol.~38, no.~16, 2024, pp. 17\,754--17\,762.

\bibitem{Singh2025}
A.~Singh, A.~Ehtesham, S.~Kumar, and T.~T. Khoei, ``Agentic retrieval-augmented generation: A survey on agentic rag,'' \emph{arXiv preprint arXiv:2501.09136}, 2025.

\bibitem{Nguyen2025}
T.~Nguyen, P.~Chin, and Y.-W. Tai, ``Ma-rag: Multi-agent retrieval-augmented generation via collaborative chain-of-thought reasoning,'' \emph{arXiv preprint arXiv:2505.20096}, 2025.

\bibitem{Pathway2025}
{Pathway Community}, ``Adaptive agents for real-time rag: Domain-specific ai for legal, finance \& healthcare,'' \url{https://pathway.com/blog/adaptive-agents-rag/}, 2025.

\bibitem{Barron2024}
R.~C. Barron, V.~Grantcharov, S.~Wanna, M.~E. Eren, M.~Bhattarai, N.~Solovyev, G.~Tompkins, C.~Nicholas, K.~. Rasmussen, C.~Matuszek, and B.~S. Alexandrov, ``Domain-specific retrieval-augmented generation using vector stores, knowledge graphs, and tensor factorization,'' \emph{arXiv preprint arXiv:2410.02721}, 2024.

\bibitem{Daiya2024}
H.~Daiya, ``Leveraging large language models (llms) for enhanced risk monitoring in fintech,'' \emph{IEEE Computer Society Tech News}, 2024.

\bibitem{Broughton2024}
M.~Broughton, ``Large language models: How they help fintechs,'' \url{https://thepaymentsassociation.org/article/large-language-models-how-they-help-fintechs/}, 2024.

\bibitem{Lumenova2024}
{Lumenova AI}, ``Ai in finance: The promise and risks of rag,'' \url{https://www.lumenova.ai/blog/ai-finance-retrieval-augmented-generation/}, 2024.

\bibitem{Leal2024}
P.~H. Leal, ``Building a financial education chatbot with retrieval-augmented generation (rag),'' \url{https://medium.com/@hlealpablo/building-a-financial-education-chatbot-with-retrieval-augmented-generation-rag-bf338aa2df09}, 2023.

\end{thebibliography}

\end{document}